\pdfoutput=1
\RequirePackage{fix-cm}
\documentclass[british]{article}

\usepackage[accepted]{icml2020} 

\usepackage[utf8]{inputenc}
\usepackage[T1]{fontenc}
\usepackage[british]{babel}
\usepackage{microtype} 

\usepackage{pifont}

\usepackage{epsfig}
\usepackage{graphicx}
\usepackage{adjustbox} 
\usepackage{xcolor}
\graphicspath{{./figures/}}
\DeclareGraphicsExtensions{.pdf,.png,.jpeg,.jpg}

\usepackage[cmex10]{amsmath} 
\usepackage{amssymb}

\usepackage{esint} 
\usepackage{commath} 
\usepackage{units} 
\usepackage{centernot} 

\usepackage{array} 

\usepackage{rotating}
\usepackage{multirow}

\usepackage{booktabs} 

\usepackage{subcaption}

\usepackage{enumitem} 

\usepackage[unicode=true,pdfusetitle,%
pdfauthor={Simon Alexanderson, Gustav Eje Henter},%
bookmarks=false,pagebackref=true,breaklinks=true,%
pdfborder={0 0 0},colorlinks,citecolor=blue]{hyperref} 

\newcommand{\bb}[1]{\boldsymbol{#1}}

\renewcommand{\mid}{\,\ifnum\currentgrouptype=16 \middle\fi|\,}

\newcommand{\real}{\mathbb{R}}

\newcommand{\Gauss}{\mathcal{N}}

\DeclareMathSymbol{\shortminus}{\mathbin}{AMSa}{"39}

\setlist{nolistsep}
\let\oldmarginpar\marginpar
\renewcommand\marginpar[1]{\-\oldmarginpar[\raggedleft\footnotesize #1]%
{\raggedright\footnotesize #1}}

\icmltitlerunning{Robust model training and generalisation with Studentising flows}

\begin{document}

\twocolumn[
\icmltitle{Robust model training and generalisation with Studentising flows}




\begin{icmlauthorlist}
\icmlauthor{Simon Alexanderson}{tmh}
\icmlauthor{Gustav Eje Henter}{tmh}
\end{icmlauthorlist}

\icmlaffiliation{tmh}{Division of Speech, Music and Hearing, KTH Royal Institute of Technology, Stockholm, Sweden} 

\icmlcorrespondingauthor{Gustav~Eje Henter}{\href{mailto:ghe@kth.se}{ghe@kth.se}}

\icmlkeywords{Statistical robustness, normalising flows, Student's t-distribution, outliers, resistant statistics, generalisation}

\vskip 0.3in
]



\printAffiliationsAndNotice{}  

\begin{abstract}
Normalising flows are tractable probabilistic models that leverage the power of deep learning to describe a wide parametric family of distributions, all while remaining trainable using maximum likelihood. We discuss how these methods can be further improved based on insights from robust (in particular, resistant) statistics. Specifically, we propose to endow flow-based models with fat-tailed latent distributions such as multivariate Student's $t$, as a simple drop-in replacement for the Gaussian distribution used by conventional normalising flows. While robustness brings many advantages, this paper explores two of them: 1) We describe how using fatter-tailed base distributions can give benefits similar to gradient clipping, but without compromising the asymptotic consistency of the method. 2) We also discuss how robust ideas lead to models with reduced generalisation gap and improved held-out data likelihood. Experiments on several different datasets confirm the efficacy of the proposed approach in both regards.
\end{abstract}

\section{Introduction}
\label{sec:intro}
Normalising flows are tractable probabilistic models that leverage the power of deep learning and invertible neural networks to describe highly flexible parametric families of distributions.
In a sense, flows combine powerful implicit data-generation architectures \citep{mohamed2016learning} of generative adversarial networks (GANs) \citep{goodfellow2014generative} with the tractable inference seen in classical probabilistic models such as mixture densities \citep{bishop1994mixture}, essentially giving the best of both worlds.

Much ongoing research into normalising flows strives to devise new invertible neural-network architectures that increase the expressive power of the flow; see \citet{papamakarios2019normalizing} for a review.
However, the invertible transformation used is not the only factor that determines the success of a normalising flow in applications.
In this paper, we instead turn our attention to the latent (a.k.a.\ 
\emph{base}) distribution that flows use.
In theory, an infinitely-powerful invertible mapping can turn any continuous distribution into any other, suggesting that the base distribution does not matter.
In practice, however, properties of the base distribution can have a decisive effect on the learned models, as this paper aims to show.
Based on insights from the field of robust statistics, we propose to replace the conventional standard-normal base distribution with distributions that have fatter tails, such as the Laplace distribution or Student's $t$.
We argue that this simple change brings several advantages, of which this paper focusses on two aspects:
\begin{enumerate}
\item \vspace{-0.5ex}It makes training more stable, providing a principled and asymptotically consistent solution to problems normally addressed by heuristics such as gradient clipping.
\item It improves generalisation capabilities of learned models, especially in cases where the training data fails to capture the full diversity of the real-world distribution.
\end{enumerate}
\vspace{-0.5ex}
We present several experiments that support these claims.
Notably, the gains from robustness evidenced in the experiments do not require that we introduce any additional learned parameters into the model.




\section{Background}
\label{sec:background}
Normalising flows are nearly exclusively trained using maximum likelihood.
We here (Sec.\ \ref{ssec:mle}) review strengths and weaknesses of that training approach; how it may suffer from low statistical robustness and how that affects typical machine-learning pipelines.
We then (Sec.\ \ref{ssec:related}) discuss prior work leveraging robust statistics for deep learning.


\subsection{Maximum likelihood and outliers}
\label{ssec:mle}
Maximum likelihood estimation (MLE) is the gold standard for parameter estimation in parametric models, both in discriminative deep learning and for many generative models such as normalising flows.
The popularity of MLE is grounded in several appealing theoretical properties.
Most importantly, MLE is consistent and asymptotically efficient under mild assumptions \cite{daniels1961asymptotic}.
Consistency means that, if the true data-generating distribution is a member of the parametric family we are using, the MLE will converge on that distribution in probability.
Asymptotic efficiency adds that, as the amount of data gets large, the statistical uncertainty in the parameter estimate will furthermore be as small as possible; no other consistent estimator can do better. 

Unfortunately, MLE can easily get into trouble in the important case of \emph{misspecified models} (when the true data distribution is not part of the parametric family we are fitting).
In particular, MLE is not always robust to \emph{outliers}%
\footnote{While many practitioners informally equate outliers with errors, the treatment in this paper is deliberately agnostic to the origin of these observations. %
After all, it does not matter whether outliers are simple errors, or represent uncommon but genuine behaviours of the data-generating process, or comprise deliberate corruptions injected by an adversary -- as long as the outlying point is in the data, its mathematical effect on our model will be the same.}:
Since $\ln 0 = -\infty$, outlying datapoints that are not explained well by a model (i.e., have near-zero probability) can have an unbounded effect on the log-likelihood and the parameter estimates found by maximising it.
As a result, MLE is sensitive to training and testing data that doesn't fit the model assumptions, and may generalise poorly in these cases.
As misspecification is ubiquitous in practical applications, many steps in traditional machine-learning and data-science pipelines can be seen as workarounds that mitigate the impact of outliers before, during, and after training.
For example, careful data gathering and cleaning to prevent and exclude idiosyncratic examples prior to training is considered best practise.
Seeing that encountering poorly explained, low-probability datapoints can lead to large gradients that destabilise minibatch optimisation, various forms of gradient clipping are commonplace in practical machine learning.
This caps the degree of influence any given example can have on the learned model.
The downside is that clipped gradient minimisation is not consistent:
Since the true optimum fit sits where the average of the loss-function gradients over the data is zero, changing these gradients means that we will converge on a different optimum in general.
Finally, since misspecification tends to inflate the entropy of MLE-fitted probabilistic models \citep{lucas2019adaptive}, it is common practice to artificially reduce the entropy of samples at synthesis time for more subjectively pleasing output; cf.\ \citet{kingma2018glow,brock2019large,henter2016minimum}.
The goal of this paper is to describe a more principled approach, rooted in robust statistics, to reducing the sensitivity to outliers in normalising flows.

\begin{figure*}[!t]
\hfill
\begin{subfigure}{.32\textwidth}
  \centering
  \includegraphics[width=.95\linewidth]{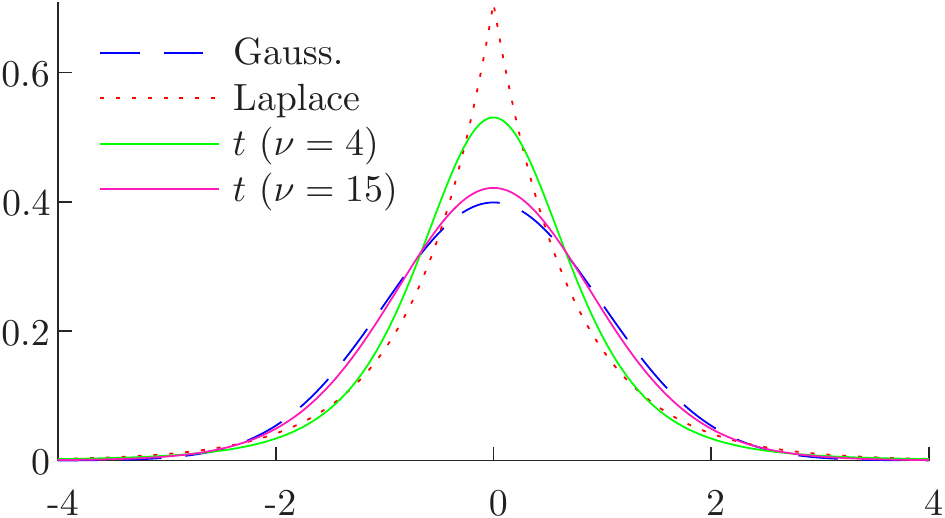}
  \caption{Probability density functions $p(x)$.}
\end{subfigure}
\hfill\hfill
\begin{subfigure}{.32\textwidth}
  \centering
  \includegraphics[width=.95\linewidth]{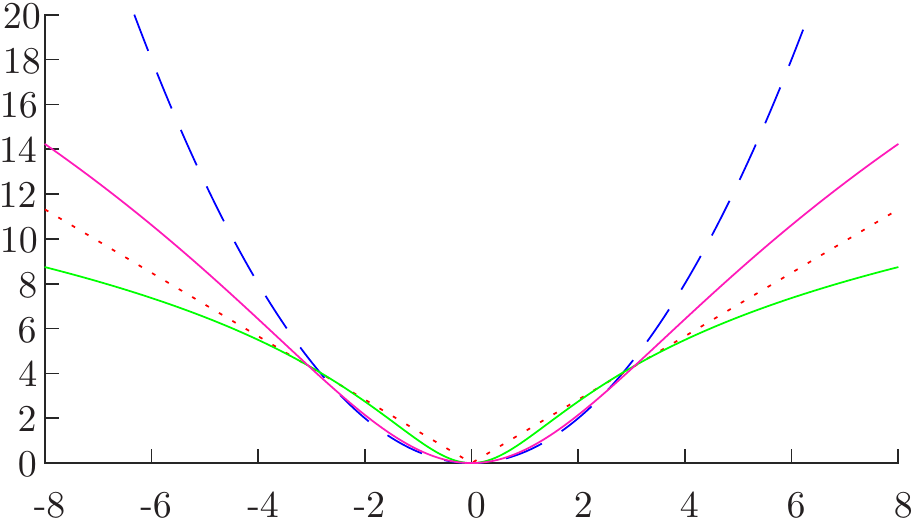}
  \caption{Penalty functions $\rho(\varepsilon)$.}
\end{subfigure}
\hfill\hfill
\begin{subfigure}{.32\textwidth}
  \centering
  \includegraphics[width=.95\linewidth]{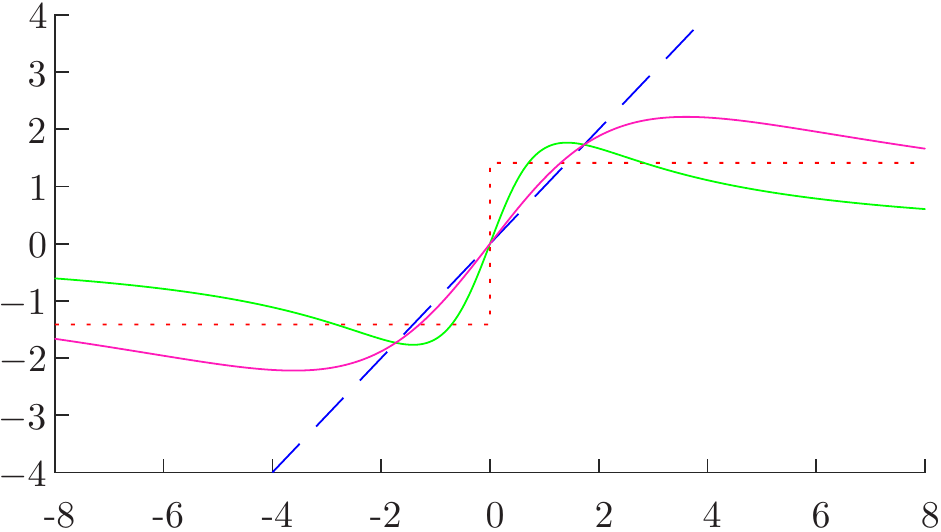}
  \caption{Influence functions $\psi(\varepsilon)$.}
  \label{sfig:influence}
\end{subfigure}
\hfill
\vspace{-0.5ex}
\caption{Functions of the normal (dashed), Laplace (dotted), and Student's $t$ distributions (solid) with mean~0 and variance~1.}
\label{fig:distributions}
\vspace{-1ex}
\end{figure*}

\subsection{Prior work}
\label{ssec:related}
Robust statistics, and in particular influence functions (Sec.\ \ref{sec:method}) have seen a number of different uses with deep learning, such as explaining neural network decisions \citep{koh2017understanding}
and subsampling large datasets \citep{ting2018optimal}.
In this work, however, we specifically consider statistical robustness in learning probabilistic models, following \citet{hampel1986robust,huber2009robust}.
This process can be made more robust in two ways: either by changing the parametric family or by changing the fitting principle.
Both the first and the second approach have been used in deep learning before.
Generative adversarial networks have been adapted to minimise a variety of divergence measures between the model and data distributions \citep{nowozin2016fgan,arjovsky2017wasserstein}, some of which amount to statistically-robust fitting principles, but they are notoriously fickle to train in practice \citep{lucic2018gans}.
\citet{henter2016robust} instead proposed using the $\beta$-divergence to fit models used in speech synthesis, demonstrating a large improvement when training on found data.
This approach does not require the use of an adversary.
However, the general idea of changing the fitting principle is unattractive with normalising flows, since maximum likelihood is the only strictly proper \emph{local} scoring function \citep[p.\ 15]{huszar2013scoring}.
This essentially means that all consistent estimation methods not based on MLE take the form of integrals over the observation space.
Such integrals are intractable to compute with the normalising flows commonly used today.

The contribution of this paper is instead to robustify flow-based models by changing the parametric family of the distributions we fit to have fatter tails than the conventional Gaussians.
Since we still use maximum likelihood for estimation, consistency is assured.
This approach has been used to solve inverse problems in stochastic optimisation \citep{aravkin2012robust} and to improve the quality of Google's production text-to-speech systems \citep{zen2016fast}.
Recently, \citet{jaini2019tails} showed that nearly all conventional normalising flows with a Gaussian base are unsuitable for modelling inherently heavy-tailed distributions.
However, they do not consider the greater advantages of changing the tail probabilities of the base distribution through the lens of robustness, which extend to data that (like much of the data in our experiments) need not have fat or heavy tails.

While there
are flow-based models with non-Gaussian base distributions such as uniform distributions \citep{muller2019neural} or GMMs
\citep{izmailov2020semi,atanov2019semi}, these do not have fat tails.
To the best of our knowledge, our work represents the first practical exploration of statistical robustness with fat-tailed distributions in normalising flows.

%

\section{Method}
\label{sec:method}
This section provides a mathematical analysis of MLE robustness, leading into our proposed solution in Sec.\ \ref{ssec:proposed}.

Our overarching goal is to mitigate the impact of outliers in training and test data using robust statistics.
We specifically choose to focus on the notion of \emph{resistant statistics}, which are estimators that do not break down under adversarial perturbation of a fraction of the data (arbitrary corruptions only have a bounded effect).
For example, among methods for estimating location parameters of distributions, the sample mean is not resistant:
By adversarially replacing just a single datapoint in the sample mean, we can make the estimator equal any value we want and make its norm go to infinity.
The median, in contrast, is resistant to up to $50$\% of the data being corrupted.

Informally, being resistant means that we allow the model to ``give up'' on explaining certain examples, in order to better fit the remainder of the data.
This behaviour can be understood through \emph{influence functions} \citep{hampel1986robust}.
In the special case of maximum-likelihood estimation of location parameters $\bb{\mu}$,
we first define the \emph{penalty function} $\rho\left(\bb{\varepsilon}\right)$ as the negative log-likelihood (NLL) loss as a function of $\bb{\varepsilon}=\bb{x}-\bb{\mu}$, offset vertically such that $\rho\left(\bb{0}\right)=0$.
The influence function $\bb{\psi}\left(\bb{\varepsilon}\right)$ is then just the gradient of $\rho$ with respect to $\bb{\varepsilon}$.
Fig.\ \ref{fig:distributions} graphs a number of different distributions in 1D, along with the associated penalty and influence functions.
For the Gaussian distribution with fixed scale, the penalty function is the squared error.
The resulting $\bb{\psi}\left(\bb{x}\right)$ is a linear function of $\bb{x}$, as plotted in Fig.\ \ref{sfig:influence}, meaning that the extent of the influence of any single outlying datapoint can grow arbitrarily large -- the estimator is not resistant.
Consequently, using maximum likelihood to fit distributions with Gaussian tails is not statistically robust.

The impact of outliers can be reduced by fitting probability distributions with fatter tails.
One example is the Laplace distribution, whose density decays exponentially with the distance from the midpoint $\bb{\mu}$; see Fig.\ \ref{fig:distributions} for plots.
The associated penalty is the absolute error $\rho\left(\bb{\varepsilon}\right)=\lVert\bb{\varepsilon}\rVert_2$.
This is minimised by the median, which is resistant to adversarial corruptions.
The Laplacian influence function in the figure is seen to be a step function and thus remains bounded everywhere, confirming that the median is resistant.
This is similar to the effect of gradient clipping in that the influence of outliers can never exceed a certain maximal magnitude.

\subsection{Proposed solution}
\label{ssec:proposed}
Define a \emph{flow} as a parametric family of densities $\{\bb{X}=\bb{f}_{\bb{\theta}}\left(\bb{Z}\right)\}_{\bb{\theta}}$, where $\bb{f}_{\bb{\theta}}$ is an invertible transformation that depends on the \emph{parameters} $\bb{\theta}\in\bb{\Theta}$ and $\bb{Z}$ is a fixed base distribution.
Our general proposal is to gain statistical robustness in this model by replacing the traditional multivariate normal base distribution by a distribution with a bounded influence function.
Our specific proposal (studied in detail in our experiments) is to replace $\bb{Z}$ by a multivariate $t$-distribution, $t_{\nu}\left(\bb{\mu},\,\bb{\Sigma}\right)$, building on \citet{lange1989robust}.
The use of multivariate $t$-distributions in flows was studied theoretically but not empirically by \citet{jaini2019tails}%
\footnote{Recent follow-up work by \citet{jaini2020tails}, appearing concurrently with our paper, does contain empirical studies of the effect of $t_{\nu}$ base distributions on the tail properties of flows.}
for the special case of triangular flows on inherently heavy-tailed data.

The pdf of the $t_{\nu}$-distribution in $D$ dimensions is
\begin{multline}
p_{t}\left(\bb{x};\,\bb{\mu},\,\bb{\Sigma},\,\nu\right)
= \Gamma\left(\tfrac{\nu+D}{2}\right) \left(\Gamma\left(\tfrac{\nu}{2}\right)\right)^{-1}
\left|\nu\pi\bb{\Sigma}\right|^{-\frac{1}{2}}\\
\cdot\left(1+\tfrac{1}{\nu}\left(\bb{x}-\bb{\mu}\right)^{\intercal}\bb{\Sigma}^{-1}\left(\bb{x}-\bb{\mu}\right)\right)^{-\frac{\nu+D}{2}}
\text{,}
\label{eq:tpdf}
\end{multline}
where the scalar $\nu>0$ is called the \emph{degrees of freedom}.
We see in Fig.\ \ref{fig:distributions} that this leads to a nonconvex penalty function and, importantly, to an influence function that approaches zero for large deviations.
This is known as a \emph{redescending} influence function, and means that outliers not only have a bounded impact in general (like for the absolute error or gradient clipping), but that gross outliers furthermore will be effectively ignored by the model.
Since the density asymptotically decays polynomially (i.e., slower than exponentially), we say that it has \emph{fat tails}.
Seeing that the (inverse) transformation $\bb{f}_{\bb{\theta}}^{-1}$ now no longer turns the observation distribution $\bb{X}$ into a normal (Gaussian) distribution, we propose to call these models \emph{Studentising flows}.

As our proposal is based on MLE, we retain both consistency and efficiency in the absence of misspecification.
In the face of outlying observations, our approach degrades gracefully, in contrast to distributions having, e.g., Gaussian tails.
As we only change the base distribution, our proposal can be combined with any invertible transformation, network architecture, and optimiser to model distributions on $\real^D$.
It can also be used with conditional invertible transformations
in order to describe conditional probability distributions.
Since the tails of $t_{\nu}\left(\bb{\mu},\,\bb{\Sigma}\right)$ get slimmer as $\nu$ increases, we can tune the degree of robustness of the approach by changing this parameter of the distribution.%
\footnote{It is also possible to treat $\nu$ as a learnable model parameter rather than a fixed or hand-tuned hyperparameter, but this procedure is not theoretically robust to gross outliers \citep{lucas1997robustness}.}
In fact, the distribution converges on the multivariate normal $\Gauss\left(\bb{\mu},\,\bb{\Sigma}\right)$ in the limit $\nu\to\infty$.
Sampling from the $t_{\nu}$-distribution can be done by drawing a sample from a multivariate Gaussian and then scaling it on the basis of a sample from the scalar $\chi^2_{\nu}$-distribution; see \citet{kotz2004multivariate}.


\begin{figure}[!t]
\centering
\includegraphics[width=0.95\columnwidth]{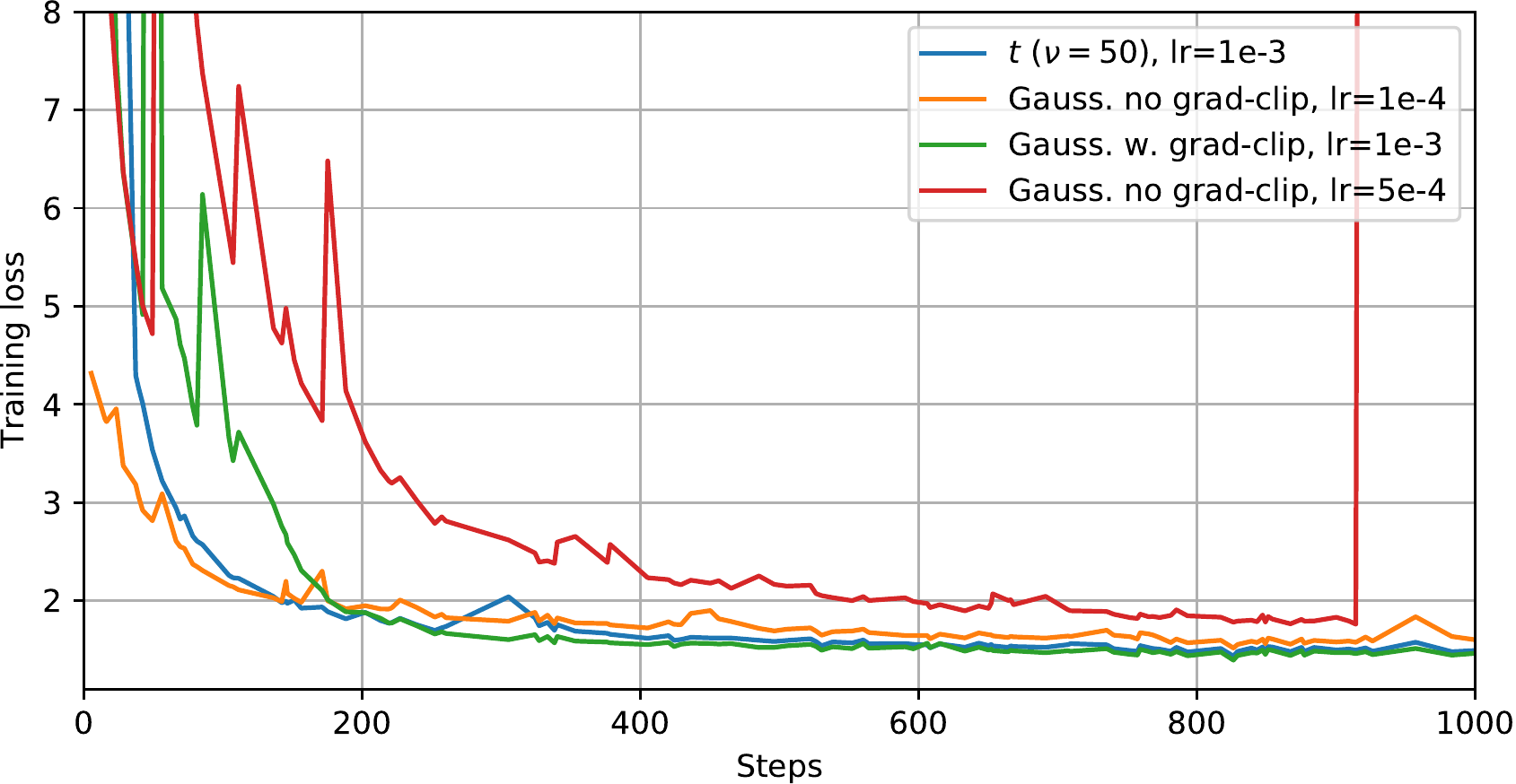}
\vspace{-0.5ex}
\caption{Training loss (NLL) on CelebA data.}
\label{fig:celeba}
\vspace{-2.9ex}
\end{figure}

\section{Experiments}
\label{sec:experiments}
In this section we demonstrate empirically the advantages of fat-tailed base distributions in normalising flows, both in terms of stable training and for improved generalisation.


\subsection{Experiments on image data}
\label{ssec:images}
Our initial experiments considered unconditional models of (uniformly dequantised) image data using Glow \citep{kingma2018glow}.
Specifically, we used the benchmark code from \citet{durkan2019neural} trained using Adam \citep{kingma2015adam}.
Implementing $t_{\nu}$-distributions for the base required just 20 lines of Python code; see Appendix \ref{sec:python}.

First we investigated training stability on the CelebA faces dataset \citep{liu2015faceattributes}.
We used the benchmark distributed by \citet{durkan2019neural}, which considers $64\times64$ images to reduce computational demands.
Our model and training hyperparameters were closely based on those used in the Glow paper,
setting $K=32$ and $L=3$ like for the smaller architectures in the article.
We found that without gradient clipping, training Glow on CelebA required low learning rates to remain stable.
As seen in Fig.\ \ref{fig:celeba}, training with learning rate $\mathrm{lr}=10^{-4}$ was stable, but training with higher learning rates $\mathrm{lr}\geq0.5\cdot10^{-3}$ did not converge.
Clipping the gradient norm at 100,
or our more principled approach of changing the base to a multivariate $t_{\nu}$-distribution (with $\nu=50$),
both enabled successful training at $\mathrm{lr}=10^{-3}$.
We also reached better log-likelihoods on held-out data than the model trained with low learning rate (see Fig.\ \ref{fig:celebaval} in Appendix \ref{sec:more}), even though the primary goal of this experiment was not necessarily to demonstrate better generalisation.
%
\begin{table}[t!]
\caption{Test-set NLL losses on MNIST with and without outliers inserted from CIFAR-10.
$\Delta$-values are w.r.t.\ the corresponding Gaussian alternative ($\nu=\infty$).
}
\label{tab:mnist}
\vspace{-0.6ex}
\centering
\small{%
\begin{tabular}{@{}c|c|cccc|cccc@{}}
\toprule
\multicolumn{1}{c}{} & Test & \multicolumn{4}{c|}{Clean} & \multicolumn{4}{c}{1\% outliers}\tabularnewline
\cline{2-10} 
Train & $\nu=$ & $\infty$ & 20 & 50 & 1000 & $\infty$ & 20 & 50 & 1000\tabularnewline
\midrule
\multirow{2}{*}{Clean} & NLL &  1.16 &  $\hphantom{\shortminus}$1.13 & $\hphantom{\shortminus}$1.13 & $\hphantom{\shortminus}$1.17 & 1.63 &  $\hphantom{\shortminus}$1.27 & $\hphantom{\shortminus}$1.26 & $\hphantom{\shortminus}$1.31\tabularnewline
 & $\Delta$ & 0 & $\shortminus$0.03 & $\shortminus$0.03 & $\hphantom{\shortminus}$0.01 & 0 & $\shortminus$0.36 & $\shortminus$0.37 & $\shortminus$0.32\tabularnewline
\midrule
1\% & NLL &  1.17 & $\hphantom{\shortminus}$1.13 & $\hphantom{\shortminus}$1.14 & $\hphantom{\shortminus}$1.18 & 1.21 & $\hphantom{\shortminus}$1.18 & $\hphantom{\shortminus}$1.19 & $\hphantom{\shortminus}$1.22\tabularnewline
outliers & $\Delta$ & 0 & $\shortminus$0.04 & $\shortminus$0.03 & $\hphantom{\shortminus}$0.01 & 0 & $\shortminus$0.03 & $\shortminus$0.02 & $\hphantom{\shortminus}$0.01\tabularnewline
\bottomrule
\end{tabular}}%
\vspace{-2.8ex}
\end{table}



Next we performed experiments on the widely-used MINST dataset \citep{lecun1998mnist} to investigate the effect of outliers on generalisation.
Since pixel intensities are bounded, image data in general does not have asymptotically fat tails.
But while MNIST is considered a quite clean dataset, we can deliberately corrupt training and/or test data by inserting greyscale-converted examples from CIFAR-10 \citep{krizhevsky2009learning}, which contains natural images that are much more diverse than the handwritten digits of MNIST.
We randomly partitioned MNIST into training, validation, and test sets (80/10/10 split), and considered four combinations of either clean or corrupted (1\% CIFAR) test and/or train+val data.
We trained (60k steps) and tested normalising as well as Studentising flows on the four combinations, using the the same learning rate schedule (cosine decay from $\mathrm{lr}=4\cdot10^{-4}$ to $10^{-4}$) and hyperparameters ($K=10$, $L=3$), and clipping the gradients for the normalising flows only.
This produced the negative log-likelihood values listed in Table \ref{tab:mnist}.
We see that, for each configuration, the proposed method performed similar to or better than the conventional setup using Gaussian base distributions.
The generalisation behaviour of $t_{\nu}$-distributions was not sensitive to the parameter $\nu$, although very high values ($\nu\approx1000$ or more) behaved more like the conventional normalising flow, as expected.
While in most cases the improvements were relatively minor, Studentising flows generalised much better to the case where the test data displayed behaviours not seen during training.

\subsection{Experiments on motion modelling}
\label{ssec:motion}
Last, we studied a domain where normalising flows constitute the current state of the art, namely conditional probabilistic motion modelling as in \citet{henter2019moglow,alexanderson2020style}.
These models resemble the VideoFlow model of \citet{kumar2020videoflow}, but also include recurrence and an external control signal.
The models give compelling visual results, but have been found to overfit significantly in terms of the log-likelihood on held-out data.
This reflects a well-known disagreement between likelihood and subjective impressions; see, e.g., \citet{theis2016note}:
Humans are much more sensitive to the presence of unnatural output examples than they are to \emph{mode dropping}, where models do not represent all possibilities the data can show.
Non-robust approaches (which cannot ``give up'' on explaining even a single observation), on the other hand, suffer significant likelihood penalties upon encountering unexpected examples in held-out data; cf.\ Table \ref{tab:mnist}.
Having methods where generalisation performance better reflects subjective output quality would be beneficial, e.g., when tuning generative models.
%
\begin{figure}[!t]
  \centering
  \includegraphics[width=0.95\columnwidth]{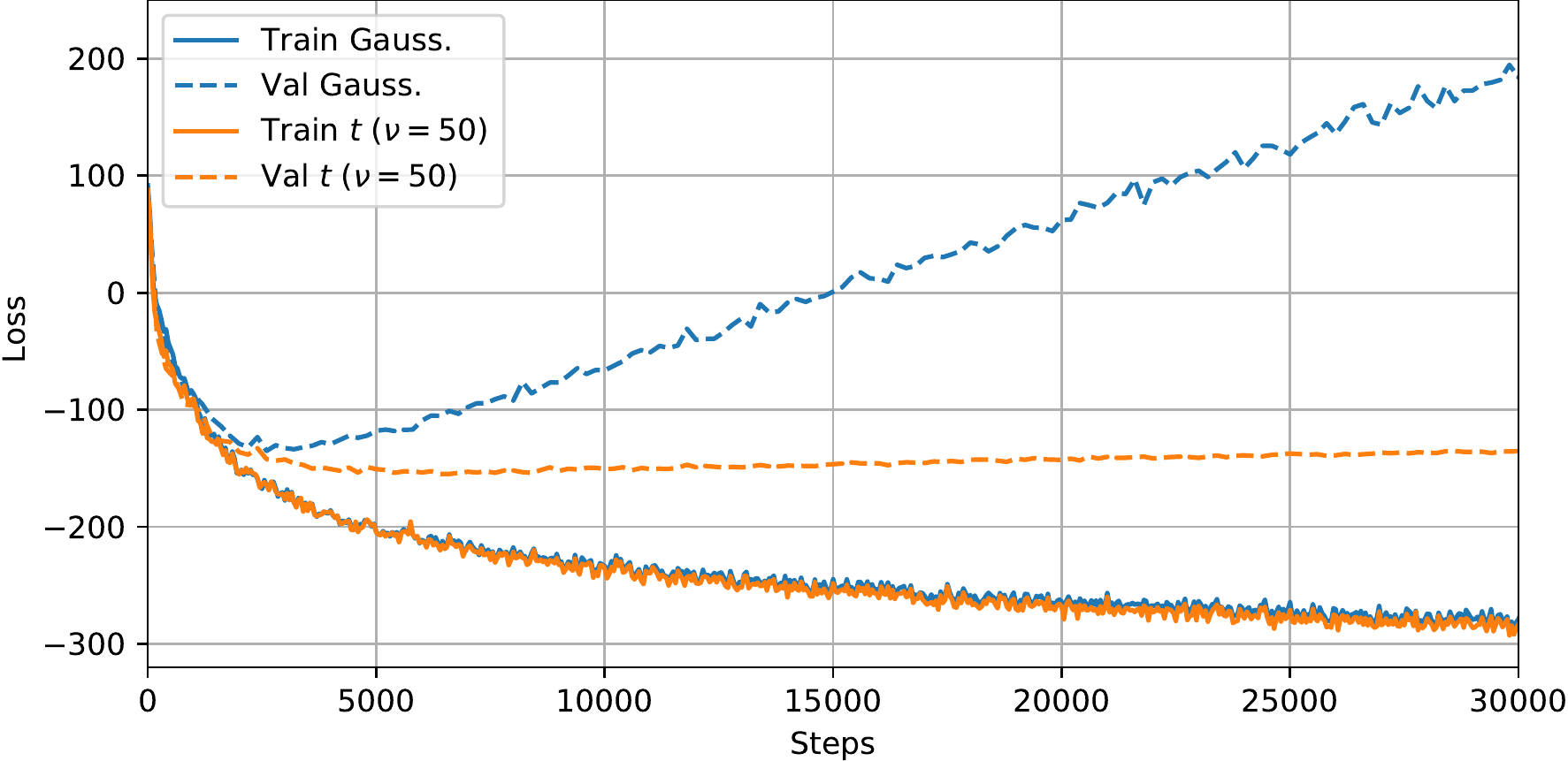}
  \vspace{-0.5ex}
  \caption{Training and validation losses on locomotion data.}
  \label{fig:locomotion}
\end{figure}
\begin{figure}[!t]
  \centering
  \includegraphics[width=0.95\columnwidth]{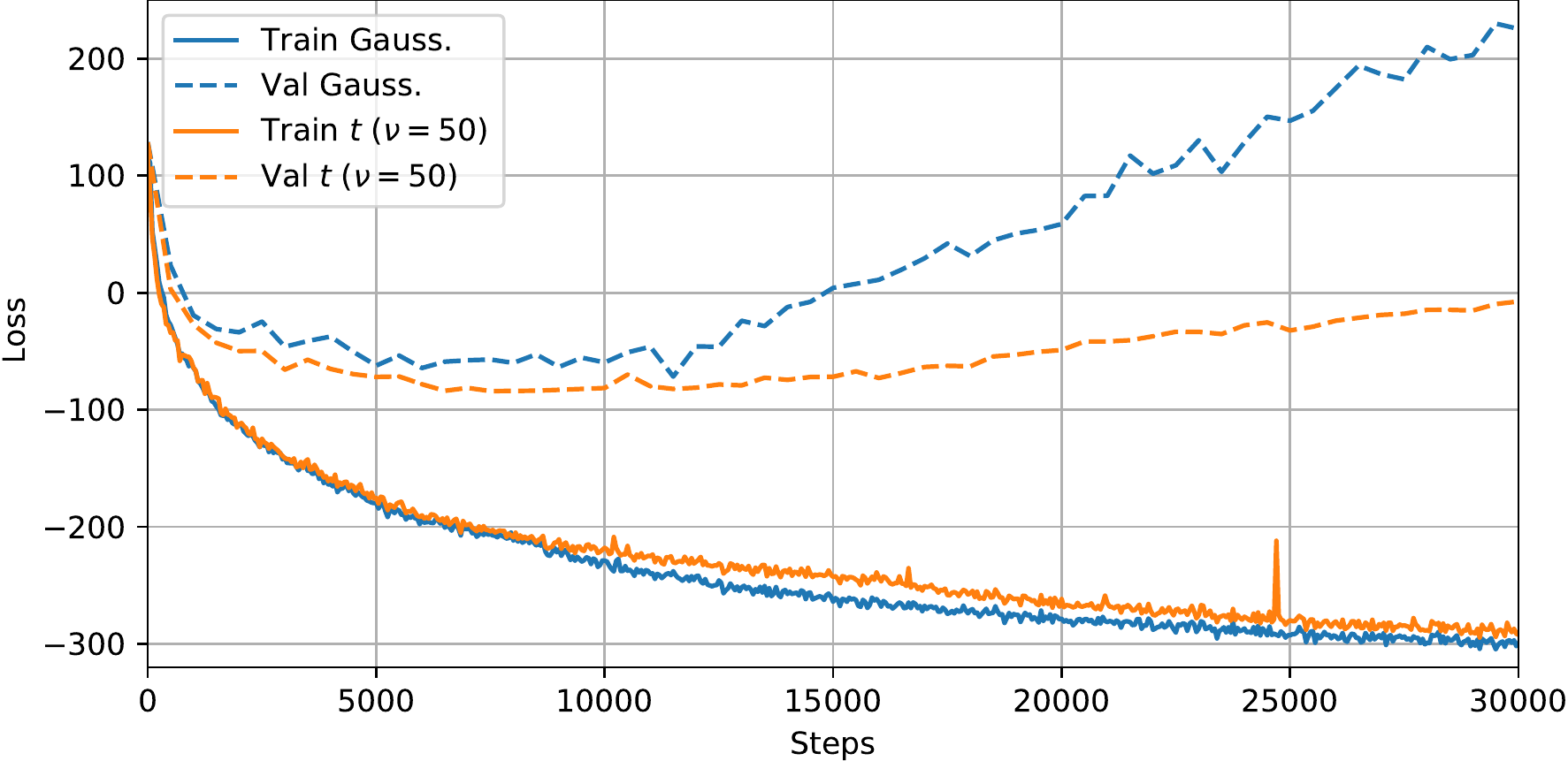}
  \vspace{-0.5ex}
  \caption{Training and validation losses on gesture data.}
  \label{fig:gesture}
  \vspace{-2ex}
\end{figure}

We considered two tasks: locomotion generation with path-based control and speech-driven gesture generation.
For locomotion synthesis, the input is a sequence of delta translations and headings specifying a motion path along the ground, and the output is a sequence of body poses (3D joint positions) that animate human locomotion along that path.
For gesture synthesis, the input is a sequence of speech-derived acoustic features and the output is a sequence of body poses (joint angles) of a character gesticulating and changing stance to the speech.
In both cases, the aim is to use motion-capture data to learn to animate plausible motion that agrees with the input signal.
See Appendix \ref{sec:more} for still images and additional information about the data.

For the gesture task we used the same model and parameters as system FB-U in \citet{alexanderson2020style}.
For the locomotion task, we found that additional tuning of the MG model from \citet{henter2019moglow} could maintain the same visual quality while reducing training time and improving performance on held-out data.
Specifically, we applied a Noam learning-rate scheduler \citep{vaswani2017attention} with peak $\mathrm{lr}=10^{-3}$ decaying to $10^{-4}$, set data dropout to 0.75, and changed the recurrent network from an LSTM to a GRU.
Learning curves for the two tasks are illustrated in Fig.\ \ref{fig:locomotion} and show similar trends.
Under a Gaussian base distribution, the loss on training data decreases, while the NLL on held-out data begins to rise steeply early on during training.%
\footnote{We have been able to replicate similarly-shaped learning curves on CelebA by changing the balance to 20\% training data and 80\% validation data (see Fig.\ \ref{fig:smalldata} in Appendix \ref{sec:more}), suggesting that the root cause of this divergent behaviour is an amount of training data that is too small to adequately sample the full diversity of natural behaviour, leading to a poor model of held-out material.
This is despite the fact that the motion databases used for these experiments are among the largest currently available for public use.
In classification, \citet{recht2019imagenet} recently highlighted similar issues of poor generalisation on new data from the same source.
}
This is subjectively misleading, since the perceived quality of randomly-sampled output motion generally keeps improving throughout training.
We note that these normalising flows were trained with gradient clipping (both of the norm and individual elements), and the smooth shape of the curves around the local optimum makes it clear that training instability is not a factor in the poor performance.

Using the same models and training setups but with our proposed $t_{\nu}$-distribution ($\nu=50$) for the base has essentially no effect on the training loss but brings the validation curves much closer to the training curves.
It is also significantly less in disagreement with subjective impressions of the quality of random motion samples with held out control-inputs.
While these plots only show the first 30k training steps, the same trends continue over the full 80k+ steps we trained, with normalising flows diverging linearly while the validation losses of Studentising flows quickly saturate; see Fig.\ \ref{fig:extended} in Appendix \ref{sec:more}.

\section{Conclusion}
\label{sec:conclusion}
We have proposed fattening the tails of the base (latent) distributions in flow-based models.
This leads to a modelling approach that is statistically robust: it remains consistent and efficient in the absence of model misspecification while degrading gracefully when data and model do not match.
We have argued that many heuristic steps in standard machine-learning pipelines, including the practice of gradient clipping during optimisation, can be seen as workarounds for core modelling approaches that lack robustness.
Our experimental results demonstrate that changing to a fat-tailed base distribution 1) provides a principled way to stabilise training, similar to what gradient clipping does, and 2) improves generalisation, both by reducing the mismatch between training and validation loss and by improving the log-likelihood of held-out data in absolute terms.
These improvements are observed for well-tuned models on datasets both with and without obviously extreme observations.
We expect the improvements due to increased robustness to be of interest to practitioners in a wide range of applications.


\section*{Acknowledgement}
This work was supported by the Swedish Research Council proj.\ 2018-05409 (StyleBot) and by the Wallenberg AI, Autonomous Systems and Software Program (WASP) of the Knut and Alice Wallenberg Foundation, Sweden.

\bibliography{refs}
\bibliographystyle{icml2020}

\clearpage
\appendix
\section{Additional information on data and results}
\label{sec:more}
\begin{figure}[!t]
  \centering
  \includegraphics[width=0.95\columnwidth]{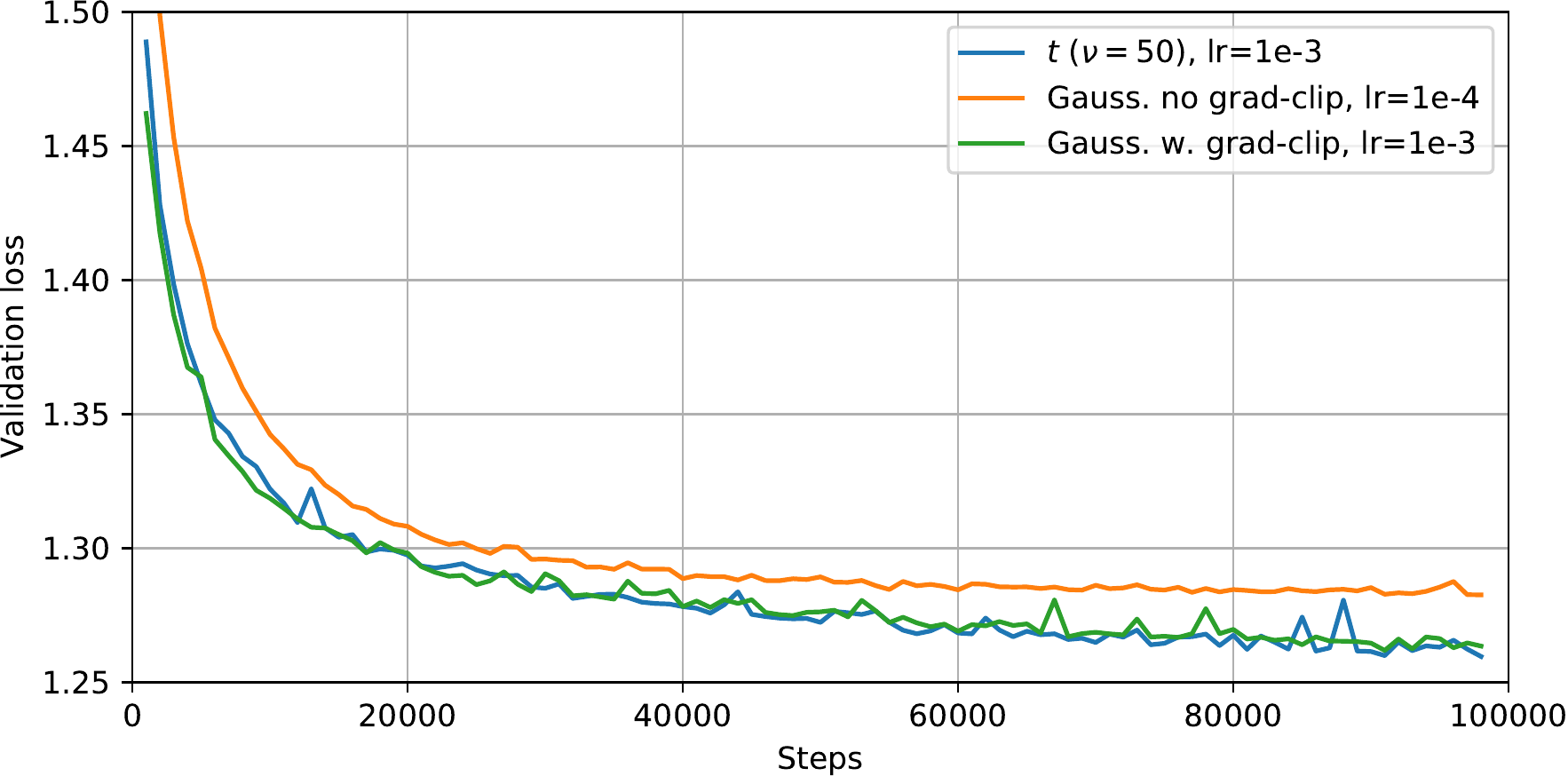}
  \vspace{-0.5ex}
  \caption{Validation loss on CelebA with different setups.}
  \label{fig:celebaval}
\end{figure}
\begin{figure}[!t]
  \centering
  \includegraphics[width=0.95\columnwidth]{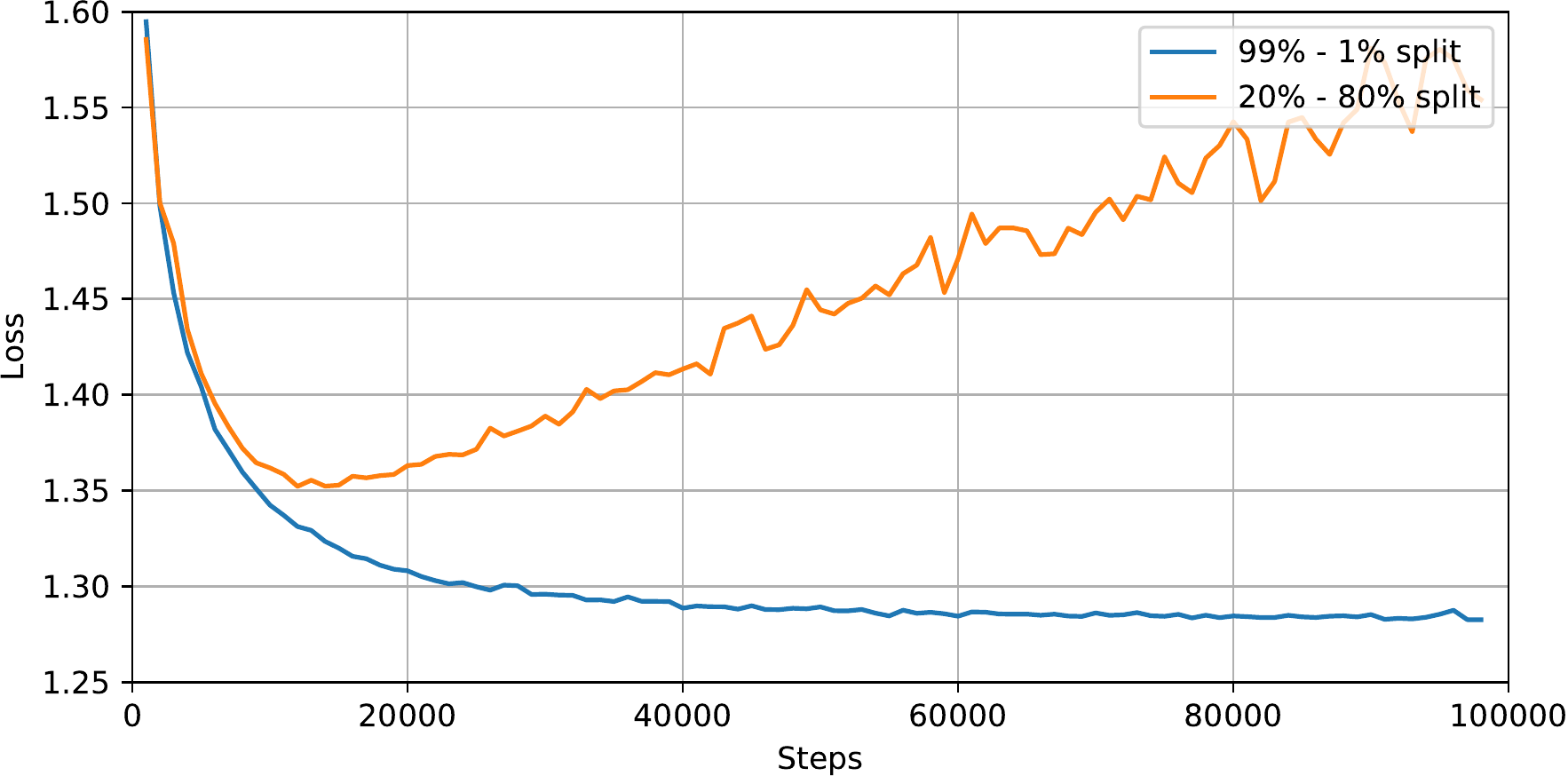}
  \vspace{-0.5ex}
  \caption{Validation loss on CelebA with different splits.}
  \label{fig:smalldata}
  \vspace{-2ex}
\end{figure}
Fig.\ \ref{fig:celebaval} reports the validation-set performance over 100k steps of training for the three stable systems from Fig.\ \ref{fig:celeba}.
We see that systems trained with the higher learning rate gave noticeably better generalisation performance.

We also performed an experiment on CelebA to see the effect of reduced training-data size on generalisation.
In particular, we tried making the training set significantly smaller than before (going from 99\% to 20\% of the database), while making the validation set much larger (from 1\% to 80\% of the database) in order to well sample the full diversity of the material.
Fig.\ \ref{fig:smalldata} shows learning curves on the CelebA data with Gaussian base distributions before and after shifting the balance between training and held-out data.
We see that, while validation loss originally decreased monotonically, the loss after changing dataset sizes instead reaches an optimum early on in the training and then begins to rise significantly again, reminiscent of the validation curves seen in Sec.\ \ref{ssec:motion}.
We conclude that the unusually large generalisation gap on the motion data at least in part can be attributed to the size of the database relative to the complexity of the task.

\begin{figure}[!t]
\begin{subfigure}{.575\columnwidth}
  \centering
  \includegraphics[width=.95\linewidth]{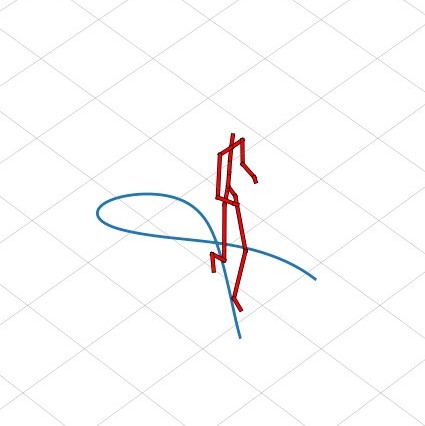}
  \caption{Locomotion with control path.}
\end{subfigure}
\hfill
\begin{subfigure}{.415\columnwidth}
  \centering
  \includegraphics[width=.95\linewidth]{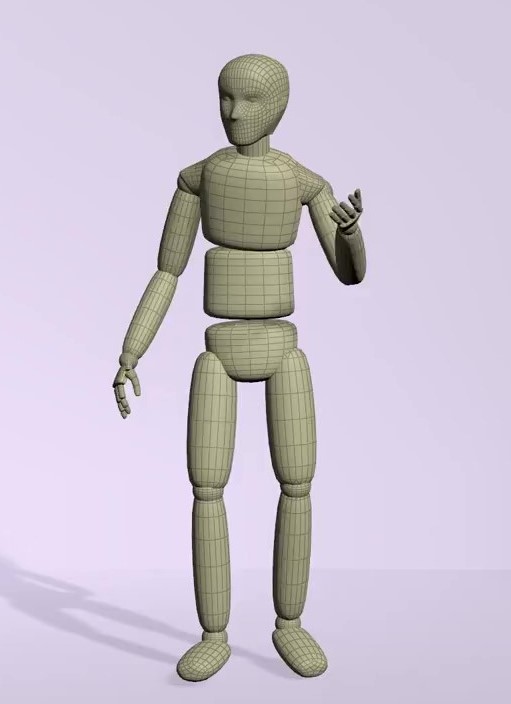}
  \caption{Gesticulating avatar.}
\end{subfigure}
\vspace{-0.5ex}
\caption{Snapshots visualising the motion data used.}
\label{fig:snapshots}
\end{figure}

\begin{figure}[!t]
\begin{subfigure}{.49\columnwidth}
  \centering
  \includegraphics[width=\linewidth]{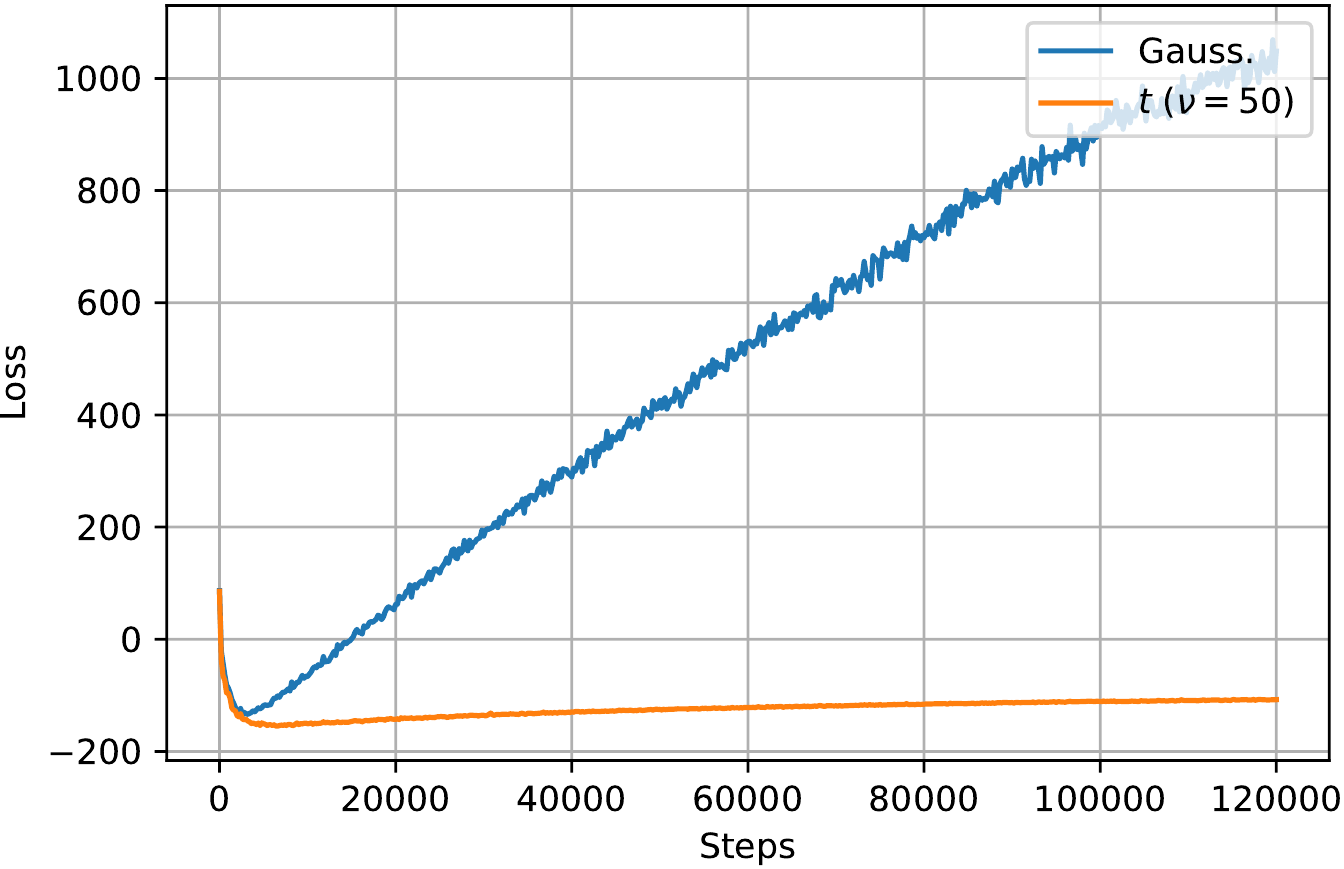}
  \caption{Locomotion modelling task.}
\end{subfigure}
\hfill
\begin{subfigure}{.49\columnwidth}
  \centering
  \includegraphics[width=\linewidth]{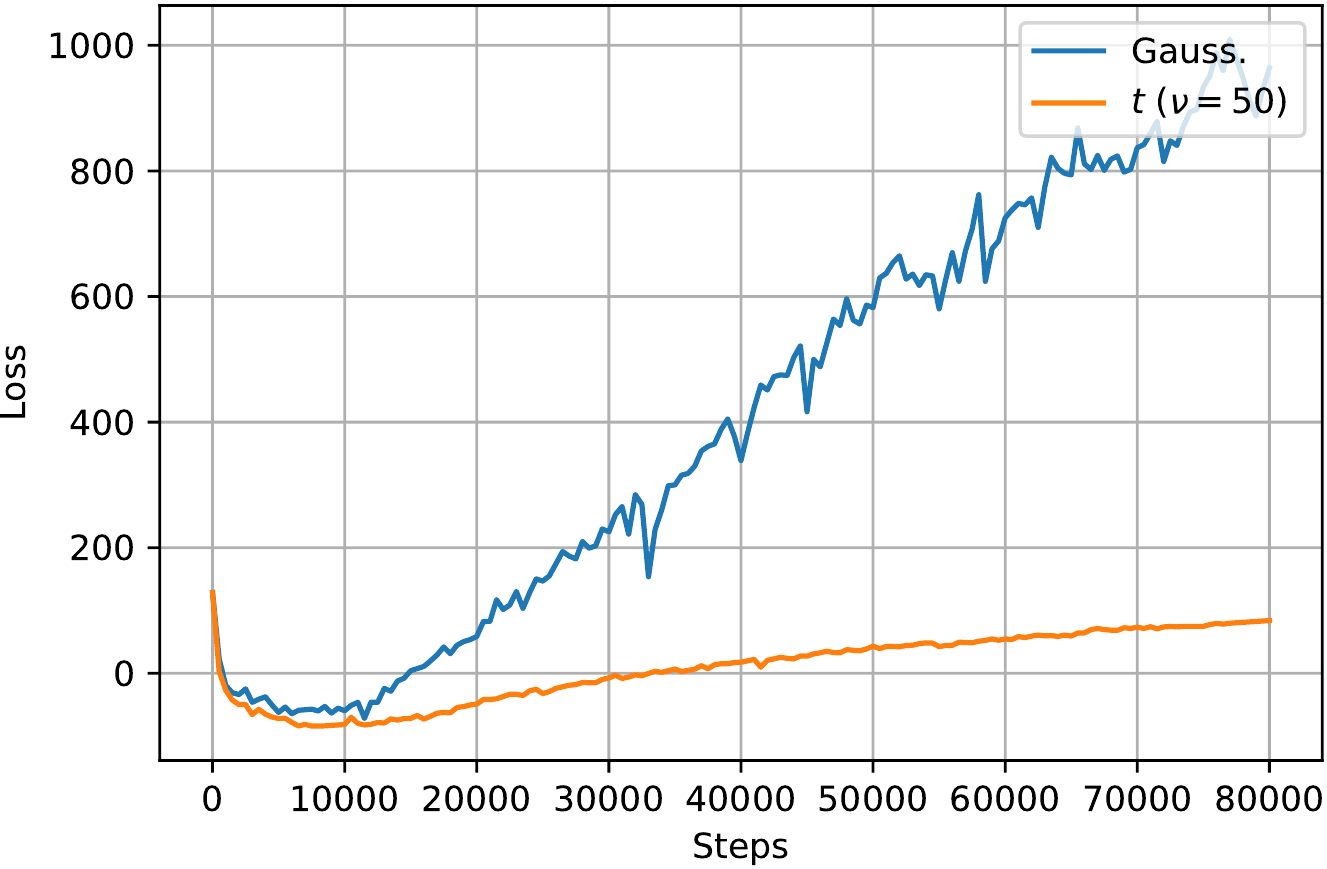}
  \caption{Gesture modelling task.}
\end{subfigure}
\vspace{-0.5ex}
\caption{Validation-loss curves of extended training.}
\label{fig:extended}
\vspace{-2ex}
\end{figure}

The two motion-data modelling tasks we considered in Sec.\ \ref{ssec:motion}, namely path-based locomotion control and speech-driven gesture generation, have applications in areas such as animation, computer games, embodied agents, and social robots.
For the locomotion data, we used the Edinburgh locomotion MOCAP database \citep{habibie2017recurrent} pooled with the locomotion trials from the trials from the CMU \citep{cmu2003mocap}
and HDM05 \citep{muller2007documentation} motion-capture databases.
Each frame in the data had an output dimensionality of $D=63$.
Gesture-generation models, meanwhile, were trained on the Trinity Gesture Dataset
collected by \citet{ferstl2018investigating},
which is a large database of joint speech and gestures.
Each output frame had $D=65$ dimensions.
Fig.\ \ref{fig:snapshots} shows still images from representative visualisations of the two tasks.
Like for image data, the numerical range of these motion datasets is bounded in practice (e.g., by the finite length of human bones coupled with the body-centric coordinate systems used in \citet{henter2019moglow}), and the data is not known to contain any numerically extreme observations.

Fig.\ \ref{fig:extended} illustrates the point from the end of Sec.\ \ref{ssec:motion} regarding the growing gap between normalising and Studentising flows over the course of the entire training.
We see that the held-out loss of the former diverges essentially linearly, while the proposed method shows saturating behaviour.


\onecolumn
\section{PyTorch code for the $t_{\nu}$-distribution}
\label{sec:python}
We here reproduce our implementation of log-probability computation and sampling with the multivariate $t_{\nu}$-distribution:
\begin{verbatim}
import numpy as np
import scipy.special
import torch

class StudentT():
    def __init__(self, shape, nu=50):
        d = shape[0]
        self._const = scipy.special.loggamma(0.5*(nu+d)) - \
              scipy.special.loggamma(0.5*nu) - 0.5*d*np.log(np.pi*nu)
        self._shape = torch.Size(shape)
        self._nu = nu

    def _log_prob(self, inputs):
        d = self._shape[0]
        input_norms = utils.sum_except_batch(((inputs)**2), num_batch_dims=1)
        likelihood = self._const - \
            0.5*(self._nu+d)*torch.log(1+(1/self._nu)*input_norms)
        return likelihood

    def _sample(self, num_samples):
        d = self._shape[0]
        x = np.random.chisquare(self._nu, num_samples)/self._nu
        x = np.tile(x[:,None], (1,d))
        x = torch.Tensor(x.astype(np.float32))
        z = torch.randn(num_samples, *self._shape)
        return (z/torch.sqrt(x))
\end{verbatim}

\end{document}